\documentclass[runningheads]{llncs}
\usepackage{graphicx}
\usepackage{times}
\usepackage{amsmath}
\usepackage{amssymb}
\usepackage{algorithm}
\usepackage{algorithmic}
\usepackage{bm}
\usepackage{color}

\begin{document}

\title{Hard-Mining Loss based Convolutional Neural Network for Face Recognition}



\author{Yash Srivastava, Vaishnav Murali and Shiv Ram Dubey\\
\institute{Computer Vision Group,\\
Indian Institute of Information Technology, Sri City, Chittoor, Andhra Pradesh, India\\
\{srivastava.y15, murali.v15, srdubey\}@iiits.in}}


\maketitle

\begin{abstract}
Face Recognition is one of the prominent problems in the computer vision domain. Witnessing advances in deep learning, significant work has been observed in face recognition, which touched upon various parts of the recognition framework like Convolutional Neural Network (CNN), Layers, Loss functions, etc. Various loss functions such as Cross-Entropy, Angular-Softmax and ArcFace have been introduced to learn the weights of network for face recognition. However, these loss functions do not give high priority to the hard samples as compared to the easy samples. Moreover, their learning process is biased due to a number of easy examples compared to hard examples. In this paper, we address this issue by considering hard examples with more priority. In order to do so, We propose a Hard-Mining loss by increasing the loss for harder examples and decreasing the loss for easy examples. The proposed concept is generic and can be used with any existing loss function. We test the Hard-Mining loss with different losses such as Cross-Entropy, Angular-Softmax and ArcFace. The proposed Hard-Mining loss is tested over widely used Labeled Faces in the Wild (LFW) and YouTube Faces (YTF) datasets. The training is performed over CASIA-WebFace and MS-Celeb-1M datasets. We use the residual network (i.e., ResNet18) for the experimental analysis. The experimental results suggest that the performance of existing loss functions is boosted when used in the framework of the proposed Hard-Mining loss.\footnote{This paper is accepted in Fifth IAPR International Conference on Computer Vision and Image Processing (CVIP), 2020.}
\keywords{Face Recognition \and Deep Learning \and Loss Functions \and Sigmoid Function \and Hard-Mining Loss.}
\end{abstract}

\section{Introduction}
\label{introduction}
In the past few years, the face recognition task has seen a tremendous growth in terms of the robust recognition and applications in various spheres of human lives. Face Recognition has been seen with a significant usage in multiple domains like biometric-based security tools and criminal identification system among many others. Such applications of the face recognition has lead to researchers and developers to work and design face recognition systems strongly built to work in an unconstrained environment as its usage is expected to grow exponentially in the forthcoming years \cite{zhou2018survey}.

The advancements in deep learning have significantly accelerated the growth and performance of face recognition. AlexNet \cite{krizhevsky2012imagenet}, proposed by Krizhevsky et al., is marked as the birth of the Convolutional Neural Networks (CNNs) which became a revolutionary architecture developed for the task of image classification and won the ImageNet Large Scale Challenge in 2012 \cite{russakovsky2015imagenet}. Since then, many CNN based approaches have been introduced for face recognition such as, DeepFace \cite{taigman2014deepface}, DeepID2 \cite{sun2014deep}, FaceNet \cite{schroff2015facenet}, SphereFace \cite{liu2017sphereface}, and ArcFace \cite{deng2018arcface}. The CNN based approaches \cite{krizhevsky2012imagenet}, \cite{he2016deep}, \cite{LBDPCNN}, \cite{hu2018squeeze}, \cite{wang2019coarse} have shown a tremendous growth in the performance as compared to the hand-crafted features \cite{dubey2016multichannel}, \cite{chakraborti2018loop}, \cite{dubey2015local}, \cite{song2018grayscale}, \cite{dubey2014rotation}, \cite{kou2018cross}. The above growth was accompanied by the development of large-scale face datasets for training and testing the CNN based models, which majorly include CASIA-Webface \cite{yi2014learning}, MS-Celeb-1M \cite{guo2016ms}, Labeled Faces in the Wild (LFW) \cite{lfw} and YouTube Faces (YTF) \cite{ytf} among other face datasets. In this work, the CASIA-Webface and MS-Celeb-1M face datasets are used for training. However, the LFW and YTF face datasets are used for the testing.

The trend of CNN over time shows that the deep CNN architectures perform better as compared to the shallow networks. It was the motivation for the deeper architectures like GoogleNet \cite{szegedy2015going} and ResNet \cite{he2016deep}. The residual network shows that the performance of the deeper plain model is not improved because it is hard to optimize such model \cite{he2016deep}. Thus, researchers also started exploring the relevance of loss functions in optimizing the deep networks. The Cross-Entropy (i.e., Softmax) loss is very widely used for optimizing the deep learning models. Recently, the work in loss functions has been quite significant with functions like SphereFace (i.e., Angular-Softmax) \cite{liu2017sphereface} and ArcFace \cite{deng2018arcface}, specially designed for the face recognition task and have shown very promising gain in the performance. Some other existing loss functions are Marginal loss \cite{deng2017marginal}, Soft-margin softmax loss \cite{liang2017soft}, Large-margin softmax loss \cite{liu2016large}, Additive margin softmax \cite{wang2018additive}, Minimum margin loss \cite{wei2018minimum}, Cosface: Large margin cosine loss \cite{wang2018cosface}, and AdaptiveFace: Adaptive margin loss \cite{liu2019adaptiveface}.
Moreover, in another work, we have conducted a performance analysis of different loss functions and found that the ArcFace outperforms other losses \cite{srivastava2019performance}. 

A few attempts are also made to utilize the complexity of data in training such as the hardest positive pairs and hardest negative pairs are computed using margin sample mining loss by Xiao et al. \cite{xiao2017margin}; an adaptive hard sample mining strategy it used by Chen et al. \cite{chen2018improving} to pick the hard examples in the training pair images; and an auxiliary embedding is used by Smirnov et al. \cite{smirnov2018hard} to pick the hard examples in mini-batches. Note that these methods try to find out the hard examples first and then use it for training. Whereas, the proposed method gives the high priority to hard examples inherently during training based on the performance of model in that iteration.

The main drawback of above mentioned loss functions is associated with its inefficiency while modelling the hard examples which lead to mis-classification. The loss due to the more number of easy examples dominates over the loss due to the less number of hard examples. This is because while training is in progress, the number of hard examples decreases while the number of easy examples increases as network learns over iterations.
In this paper, we address the above mentioned problem by giving more importance to hard examples through loss function in each iteration. We propose the Hard-Mining loss which increases the loss for the hard samples leading to high loss and decreases the loss for the easy samples leading to low loss. As a result, the average loss contains the significant contributions from the hard examples. 

This paper is structured as follows: Section  2 proposes the Hard-Mining loss and existing losses in the Hard-Mining framework; Section 3 describes the experimental setup and details about the architecture and training and testing face datasets used. Section 4 presents the experimental results and comparisons; and finally, Section 5 concludes the paper with summarizing remarks.

\section{Proposed Hard-Mining Loss}
The loss functions are used in deep learning to judge the goodness of any model under given parameters. The stochastic gradient descent (SGD) optimization is widely adapted to train the Convolutional Neural Networks (CNNs). The SGD computes the gradient of loss function w.r.t. to the parameters which is used to update that parameter such that in the next iteration, the loss should decrease. Thus, the loss functions judge the performance of the designed architecture as well as guide the learning process. It is shown in introduction that most of the existing losses are not able to penalize the mis-classification efficiently caused by harder examples. In this paper, we propose the concept of Hard-Mining loss which increases the loss for harder examples and decreases the loss for easier examples such that the average loss should have the better representation of hard examples. A comparison between the Cross-Entropy loss and proposed Hard-Mining loss is presented in Fig. \ref{fig:lossfunction} as a function of probability of being classified in the correct class. In this section, first we present the Cross-Entropy loss, then we propose the idea of Hard-Mining loss, and finally we extend the existing losses such as Cross-Entropy, Angular-Softmax, and ArcFace in the proposed Hard-Mining framework.

\subsection{Cross-Entropy Loss}
The Cross-Entropy (or softmax) loss has been majorly used to judge the performance of CNN models for image classification task \cite{krizhevsky2012imagenet}, \cite{he2016deep}. Mathematically, the Cross-Entropy loss can be given as
\begin{equation}
\mathcal{L}_{CE}=-\frac{1}{N}\sum_{i=1}^{N}\log\frac{e^{W^T_{y_i} x_i+b_{y_i}}}{\sum_{j=1}^{n}e^{W^T_j x_i+b_j}},
\label{eq:softmax}
\end{equation}
where $W$ is the weight matrix, $b$ is the bias term, $x_i$ is the $i^{th}$ training sample, $y_i$ is the class label for $i^{th}$ training sample, $N$ is the number of samples, $W_j$ and ${W}_{y_i}$ are the $j^{th}$ and $y_i^{th}$ columns of ${W}$, respectively. The Cross-Entropy loss is used as the baseline by the recent loss functions such as Angular-Softmax and ArcFace over the face recognition problem. Hence, we also use the Cross-Entropy loss as the baseline along with Angular-Softmax and ArcFace losses.

\begin{figure}[!t]
\centering
\includegraphics[trim=15 0 20 10, clip, width=0.5\linewidth]{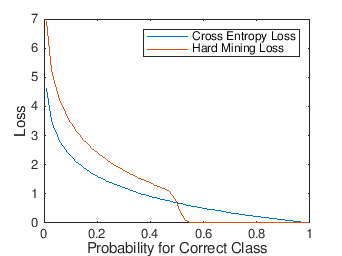}
\caption{Loss value vs Likelihood (i.e., probability for correct class) plot for the Cross-Entropy loss and Hard-Mining loss functions. Note that the Hard-Mining loss is computed on the output of Cross-Entropy loss.}
\label{fig:lossfunction}
\end{figure}

The behavior of the Cross-Entropy loss w.r.t. the probability of being classified in the correct class for an example is plotted in Fig. \ref{fig:lossfunction}. It can be observed from this analysis that the Cross-Entropy loss gradually follows a downward slope and there is no big difference between easy and hard examples. We believe that if the probability is more than $0.5$ then the loss should be minimum. Whereas, if the probability is less than $0.5$ then the loss should be on higher side. This is our intution to propose the Hard Mining Loss described next.

\subsection{Hard-Mining Loss}
Motivated from the fact that the loss for harder examples should be more, we propose the idea of Hard-Mining loss. The proposed Hard-Mining loss increases the loss if the probability is less than roughly $0.5$, while at the same time it also decreases the loss if probability is more than $0.5$ roughly. The Hard-Mining loss is defined as
\begin{equation}
\mathcal{L}_{{HM}}= \alpha \times \mathcal{L} \times \sigma(\beta \times \mathcal{L})
\label{eq:hardmining}
\end{equation}
where $\mathcal{L}$ is the loss generated by any other loss function such as Cross-Entropy, Angular-Softmax, etc., $\alpha$ and $\beta$ are the hyperparameters and $\sigma$ is the sigmoid function given as:
\begin{equation}
\mathcal{\sigma}(x)= \frac{1}{1 + e^{-A(x-B)}}
\label{eq:sigmoid}
\end{equation}
where $A$ and $B$ are the hyperparameters. 

Note that the Hard-Mining operation is generic in nature, i.e., it can be used along with any existing loss function. In this paper, we use the Hard-Mining operation along with Cross-Entropy, Angular-Softmax, and ArcFace losses.

\subsection{Hard-Mining Cross-Entropy Loss}
As mentioned previously, the Hard-Mining concept is generic and can be used with existing losses. Primarily, we define the Hard-Mining loss with Cross-Entropy loss. The Hard-Mining Cross-Entropy loss ($\mathcal{L}_{HM\_CE}$) is defined as
\begin{equation}
\mathcal{L}_{HM\_CE}= \alpha * \mathcal{L}_{CE} * \sigma(\beta * \mathcal{L}_{CE})
\label{eq:hm-ce}
\end{equation}
where $\alpha$ and $\beta$ are the hyperparameters, $\sigma$ is defined in (\ref{eq:softmax}), and $\mathcal{L}_{CE}$ is the Cross-Entropy loss given in (\ref{eq:softmax}). Algorithm \ref{alg1} shows the step-by-step instructions for the proposed Hard-Mining Cross-Entropy loss ($\mathcal{L}_{HM\_CE}$).

The behavior of Hard-Mining operation on Cross-Entropy loss is depicted in Fig. \ref{fig:lossfunction}. Note that the values of hyper-parameters $\alpha$, $\beta$, $A$, and $B$ are set to $1.5$, $1.1$, $35$, and $0.75$, respectively. It can be seen that the Hard-Mining operation increases the loss for hard examples (i.e., with less than half probability) while it decreases the loss for easy examples (i.e., with more than half probability). Our definition of hard/easy examples is relative to the probability of being classified in the correct class in a given iteration. Thus, the hard examples at the start of the training might become easy examples after training of some iterations.

\begin{algorithm}[!t]
\begin{algorithmic}
\STATE \textbf{Input:} Predicted class scores, Ground-truth class label, and hyper-parameters $\alpha$, $\beta$, $A$ and $B$.
\STATE \textbf{Output:} Loss generated.
\begin{enumerate} \setlength{\itemsep}{-\itemsep}
\item $\mathcal{L}_{CE} \leftarrow CrossEntropy(input, target, W)$
\item $x \leftarrow \beta \times \mathcal{L}_{CE}$
\item $y \leftarrow A \times (x-B)$
\item $z \leftarrow Sigmoid(y)$
\item $\mathcal{L}_{HM\_CE} \leftarrow \alpha \times x \times z$
\item return $\mathcal{L}_{HM\_CE}$
\end{enumerate}
\end{algorithmic}
\caption{Hard-Mining Cross-Entropy Loss Algorithm}
\label{alg1}
\end{algorithm}

Since, the Cross-Entropy is a very widely used loss function in various machine learning problems, it is paramount that we study the performance of Hard-Mining operation with loss functions specially designed for the face recognition problem. We consider two loss functions (i.e., Angular-Softmax \cite{liu2017sphereface} and ArcFace \cite{deng2018arcface}) designed for the face recognition problem in the proposed Hard-Mining loss framework.

\subsection{Hard-Mining Angular-Softmax Loss}
The Hard-Mining Angular-Softmax loss ($\mathcal{L}_{HM\_AS}$) is defined as follows:
\begin{equation}
\mathcal{L}_{HM\_AS}= \alpha \times \mathcal{L}_{AS} \times \sigma(\beta \times \mathcal{L}_{AS})
\label{eq:swish-loss}
\end{equation}
where $\alpha$ and $\beta$ are the hyper-parameters, $\sigma$ is given in (\ref{eq:sigmoid}), and $\mathcal{L}_{AS}$ is the Angular-Softmax loss defined in the SphereFace model \cite{liu2017sphereface} and given as
\begin{equation}
\mathcal{L}_{AS}=-\frac{1}{N}\sum_{i=1}^{N}\log\big( \frac{e^{\|\bm{x}_i\|\psi(\theta_{y_i,i})}}{e^{\|\bm{x}_i\|\psi(\theta_{y_i,i})}+
\sum_{j\neq y_i}e^{\|\bm{x}_i\|\cos(\theta_{j,i})}} \big)
\label{angular}
\end{equation}
where $x_i$ is the $i^{th}$ training sample, $\thickmuskip=2mu \medmuskip=2mu \psi(\theta_{y_i,i})=(-1)^k\cos(m\theta_{y_i,i})-2k$ for $ \theta_{y_i,i}\in[\frac{k\pi}{m},\frac{(k+1)\pi}{m}]$, $\thickmuskip=2mu k\in[0,m-1]$ and $\thickmuskip=2mu m\geq1$ is an integer controlling the size of angular margin.

\subsection{Hard-Mining ArcFace Loss}
ArcFace loss has been used in the recently developed ArcFace model for face recognition \cite{deng2018arcface}. In a recent performance comparison study, ArcFace has been figured as the outstanding loss for face recognition \cite{srivastava2019performance}. 
The Hard-Mining ArcFace loss ($\mathcal{L}_{HM\_AF}$) is defined as
\begin{equation}
\mathcal{L}_{HM\_AF}= \alpha \times \mathcal{L}_{AF} \times \sigma(\beta \times \mathcal{L}_{AF})
\label{eq:swish-lossq}
\end{equation}
where $\alpha$ and $\beta$ are the hyper-parameters, $\sigma$ is given in (\ref{eq:sigmoid}), and $\mathcal{L}_{AF}$ is the ArcFace loss \cite{deng2018arcface} and given as
\begin{equation}
\mathcal{L}_{AF} =-\frac{1}{N}\sum_{i=1}^{N}\log\frac{e^{s \cdot (\cos(\theta_{y_i}+m))}}{e^{s \cdot(\cos(\theta_{y_i}+m))}+\sum_{j=1,j\neq  y_i}^{n}e^{s \cdot \cos\theta_{j}}},
\label{eq:aml}
\end{equation}
where $s$ is the radius of the hypersphere, $m$ is the additive angular margin penalty between $x_i$ and ${W_y}_{i}$, and $\cos(\theta+m)$ is the margin which makes the class-separations more stringent. 


\section{Experimental Setup}
\label{experimentalsetup}
In this section, we discuss the CNN architectures, training and testing datasets used for the experiments along with other settings like optimizers, learning rate, epochs, etc.

\subsection{CNN Architectures}
\label{cnnarchitectures}
Several CNN architectures have been developed for different computer vision tasks. The recent trend is to utilize the power of residual learning. The ResNet model uses the residual blocks \cite{he2016deep} which is very commonly used nowadays. In this paper, we consider ResNet architecture with 18 depth (i.e., ResNet18) for all the experiments.

\subsection{Training Datasets}
\label{trainingdataset}
In our experiments, we primarily use two publicly available datasets such as CASIA-Webface \cite{yi2014learning} and MS-Celeb-1M \cite{guo2016ms} as the training datasets. The CASIA-Webface is one of the most widely adapted and available dataset used for the face recognition task. It contains 4,94,414 colored face images belonging to 10,575 different individuals. Second dataset used in our experiments is the MS-Celeb-1M dataset which consists of 1,00,000 face identities with each class containing 100 images leading to about 10M images, which are scraped from public sources. Being a humongous dataset, it contains a lot of noise and variations which impact the performance of the trained model. Hence, we use a cleaned and refined subset of the dataset as per the cleaned list provided by the ArcFace \cite{he2016deep} authors.

\subsection{Testing Datasets}
\label{testingdataset}
We use the Labeled Faces in the Wild (LFW) \cite{lfw} and Youtube Faces (YTF) \cite{ytf} as the testing datasets in this paper. The LFW dataset contains $13,233$ images of $5749$ identities. The YTF dataset consists of $3,425$ videos of $1,595$ different people with images available in frame-by-frame format and retrieved through the provided meta data. Both the datasets use the standard LFW benchmark for face verification, which provide the verification accuracies over the testing dataset. These accuracies are used as the performance measure in the state-of-the-art face recognition works. Hence, we also use the accuracy as the performance measure in this paper.

\subsection{Input Data and Network Settings}
\label{settings}
Following the recent trend \cite{deng2018arcface}, \cite{liu2017sphereface}, we use the MTCNN \cite{zhang2016joint} to align the face images. The images are normalized by subtracting 127.5 from each pixel and then being divided by 128. The batch-size is kept at $64$ with the initial learning rate as $0.01$. The learning rate is multiplied by $0.1$ at $8^{th}$, $12^{th}$ and $16^{th}$ epochs. The model is trained up to $20$ epochs. The Stochastic Gradient Descent with Momentum (SGDM) is used as the optimizer to train the network. The values of hyper-parameters $\alpha$, $\beta$, $A$, and $B$ are empirically set to $1.5$, $1.1$, $35$, and $0.75$, respectively, in this paper.

\begin{table*}[!t]
\centering
\caption{Verification accuracies (\%) using ResNet18 model over LFW and YTF face recognition testing datasets under different loss functions. The training is performed over CASIA-WebFace dataset.}
\centering
\begin{tabular}{|p{6cm}|p{3cm}|p{3cm}|}
\hline
\textbf{Loss Function} & \textbf{Accuracy on LFW Dataset} & \textbf{Accuracy on YTF Dataset} \tabularnewline
\hline
\hline
Cross-Entropy loss ($\mathcal{L}_{CE}$) & 95.35 & 91.8 \tabularnewline
\hline
Hard-Mining Cross-Entropy loss ($\mathcal{L}_{HM\_CE}$) & 96.75 &	93.1\tabularnewline
\hline
\hline
Angular-Sofmax loss ($\mathcal{L}_{AS}$) & 97.12 & 93.9\tabularnewline
\hline
Hard-Mining Angular-Sofmax loss ($\mathcal{L}_{HM\_AS}$) & 97.3 & 94.1\tabularnewline
\hline
\hline
ArcFace loss ($\mathcal{L}_{AF}$) & 97.79 & 94.54\tabularnewline
\hline
Hard-Mining ArcFace loss ($\mathcal{L}_{HM\_AF}$) & 97.9 & 94.67\tabularnewline
\hline
\end{tabular}
\label{results_casia}
\end{table*}

\section{Experimental Results and Observations}
\label{performance}
In order to show the effect of the proposed Hard-Mining loss, the face recognition experiments are conducted in this paper with ResNet18 model. Three existing loss functions, namely Cross-Entropy, Angular-Softmax and ArcFace, are used in the framework of the proposed Hard-Mining loss. The training is performed over the CASIA-WebFace and MS-Celeb-1M datasets and testing is perfomed over the LFW and YTF datasets.

\begin{table*}[!t]
\centering
\caption{Verification accuracies (\%) using ResNet18 model over LFW and YTF face recognition testing datasets under different loss functions. The training is performed over MS-Celeb-1M dataset.}
\centering
\begin{tabular}{|p{6cm}|p{3cm}|p{3cm}|}
\hline
\textbf{Loss Function} & \textbf{Accuracy on LFW Dataset} & \textbf{Accuracy on YTF Dataset} \tabularnewline
\hline
\hline
Cross-Entropy los ($\mathcal{L}_{CE}$) &	95.1 & 92.45\tabularnewline
\hline
Hard-Mining Cross-Entropy loss ($\mathcal{L}_{HM\_CE}$) &	95.1 & 92.5\tabularnewline
\hline
\hline
Angular-Softmax loss ($\mathcal{L}_{AS}$) & 96.9 & 94.1\tabularnewline
\hline
Hard-Mining Angular-Softmax loss ($\mathcal{L}_{HM\_AS}$) & 97.05 & 93.8\tabularnewline
\hline
\hline
ArcFace loss ($\mathcal{L}_{AF}$) & 97.6 & 95.1\tabularnewline
\hline
Hard-Mining ArcFace loss ($\mathcal{L}_{HM\_AF}$) & 98 & 94.9\tabularnewline
\hline
\end{tabular}
\label{results_msceleb}
\end{table*}

The results in terms of the verification accuracies are reported in Table \ref{results_casia} using ResNet18 model for the CASIA-WebFace training dataset over the LFW and YTF testing datasets. It can be seen that an improvement is obtained by the Hard-Mining Cross-Entropy loss, Hard-Mining Angular-Softmax loss, and Hard-Mining ArcFace loss as compared to the Cross-Entropy loss, Angular-Softmax loss, and ArcFace loss, respectively, over both the LFW and YTF datasets.

The results in terms of the verification accuracies are reported in Table \ref{results_msceleb} using ResNet18 model for the MS-Celeb-1M training dataset over the LFW and YTF testing datasets. It is noticed from this result that the performance of Hard-Mining operation based losses is either better or comparable over LFW dataset w.r.t. the losses without Hard-Mining operation. Moreover, Hard-Mining operation is also suited with Cross-Entropy loss over YTF dataset when training is performed over MS-Celeb-1M datasets.

The experimental results suggest that increasing the loss for harder examples and decreasing the loss for easy examples in each iteration enforce the network to learn the characteristics of hard-examples as well. Overall, the proposed Hard-Mining loss is well suited for the face recognition problem along with the existing loss functions.

\section{Conclusion}
\label{conclusion}
In this paper, a concept of Hard-Mining loss is proposed which increases the loss for hard examples being mis-classified and decreases the loss for easy examples. By doing so, we enforce the network to learn the characteristics of hard examples. The proposed concept is generic in nature and can be used with any existing loss function. We have tested the proposed Hard-Mining loss with Cross-Entropy, Angular-Softmax and ArcFace losses. The experiments are performed over CASIA-WebFace and MS-Celeb-1M training datasets and LFW and YTF testing datasets using ResNet18 model. It is observed from the experiments that the proposed Hard-Mining loss boosts the performance of existing losses in most of the cases.

\bibliographystyle{splncs04}
\bibliography{references}

\end{document}